\documentclass[runningheads]{llncs}
\usepackage[T1]{fontenc}
\usepackage{graphicx}
\usepackage{multirow}
\usepackage{booktabs}
\usepackage{amsmath}
\usepackage{amsfonts}
\usepackage{makecell}
\usepackage{hyperref}
\usepackage{tablefootnote}
\usepackage{multicol}

\usepackage{times}
\usepackage{soul}
\usepackage{url}
\usepackage[utf8]{inputenc}
\usepackage{arydshln}

\usepackage{color}

\begin{document}

\title{Enhancing Complex Causality Extraction via Improved Subtask Interaction and Knowledge Fusion}
\titlerunning{Enhancing CCE via Improved SI and KF}
% If the paper title is too long for the running head, you can set
% an abbreviated paper title here
%  
\author{Jinglong Gao \inst{1,2} \and Chen Lu \inst{2} \and Xiao Ding \inst{1*} \and Zhongyang Li \inst{3} \and Ting Liu \inst{1} \and Bing Qin \inst{1}
	}
\authorrunning{J. Gao et al.}
% First names are abbreviated in the running head.
% If there are more than two authors, 'et al.' is used.
%
\institute{Research Center for Social Computing and Information Retrieval,\\Harbin Institute of Technology, China \and
	National Key Laboratory of Information Systems Engineering, Nanjing, China \and Huawei Cloud, Shenzhen, China\\
	\email{\{jlgao, xding, tliu, qinb\}@ir.hit.edu.cn}\\
 \email{luchen3@cetc.com.cn}\\
 \email{lizhongyang6@huawei.com}}

\maketitle              % typeset the header of the contribution
\begin{abstract}
Event Causality Extraction (ECE) aims at extracting causal event pairs from texts.
Despite ChatGPT's recent success, fine-tuning small models remains the best approach for the ECE task.
However, existing fine-tuning based ECE methods cannot address all three key challenges in ECE simultaneously:
1)~\textbf{Complex Causality Extraction}, where multiple causal-effect pairs occur within a single sentence; 2)~\textbf{Subtask~Interaction}, which involves modeling the mutual dependence between the two subtasks of ECE, i.e., extracting events and identifying the causal relationship between extracted events; and 3)~\textbf{Knowledge Fusion}, which requires effectively fusing the knowledge in two modalities, i.e., the expressive pretrained language models and the structured knowledge graphs.
In this paper, we propose a unified ECE framework (\textbf{UniCE}) to address all three issues in ECE simultaneously.
Specifically, we design a subtask interaction mechanism to enable mutual interaction between the two ECE subtasks.
Besides, we design a knowledge fusion mechanism to fuse knowledge in the two modalities.
Furthermore, we employ separate decoders for each subtask to facilitate complex causality extraction.
Experiments on three benchmark datasets demonstrate that our method achieves state-of-the-art performance and outperforms ChatGPT with a margin of at least 30\% F1-score.
More importantly, our model can also be used to effectively improve the ECE performance of ChatGPT via in-context learning.
% The abstract should briefly summarize the contents of the paper in
% 150--250 words.

\end{abstract}
\renewcommand{\thefootnote}{\fnsymbol{footnote}}
\footnotetext[1]{Corresponding Author}
\section{Introduction}
\begin{figure}[t]
\centering
\includegraphics[width=0.85\linewidth]{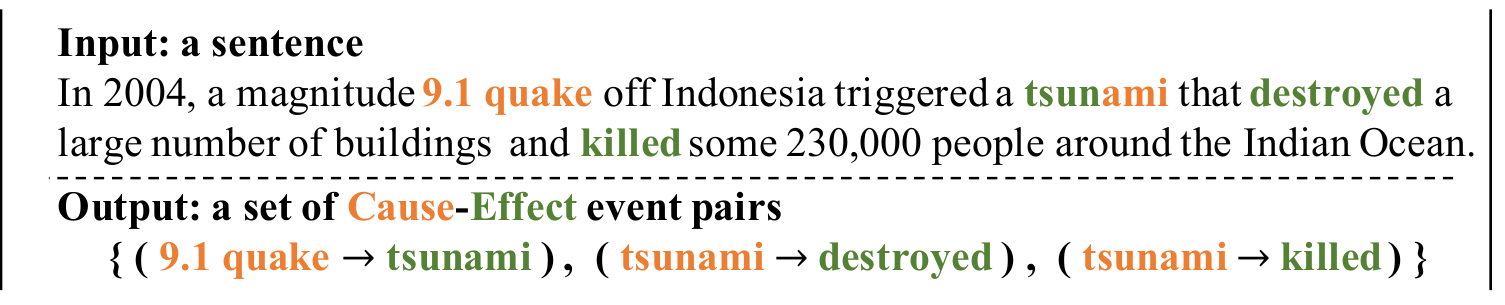}
  \caption{An example of the ECE task. All the arrows are from the cause event to the effect event.}
  \label{fig:example-of-ece}
\end{figure}

Event Causality Extraction (ECE) aims to extract causal~event~pairs from texts.
As shown in Figure~\ref{fig:example-of-ece}, given the input sentence, an ECE system should extract all cause-effect event pairs.

Most existing methods address ECE with a pipeline framework \cite{kadowaki2019event} that includes two subtasks: 1) Event Extraction (EE) \cite{fu2024jun,li2021jun}, which extracts events that may be causally related to other events in the input sentence; and 2) Event Causality Identification (ECI) \cite{liu2020knowledge}, which identifies the causal relationship between extracted events. However, the pipeline framework ignores the mutual dependence between the two subtasks.
Firstly, since only gold labeled events are used to train the ECI model, the later stage (ECI) cannot adapt to errors in the early stage (EE).
Secondly, the causal relations identified by the ECI model provide useful knowledge for the EE model to extract events.

To jointly learn EE and ECI, several previous works simplified ECE into a sequence labeling task \cite{jinghang2020causal}, which can extract one cause-effect pair in a single sentence.
However, these methods struggle to handle sentences containing multiple cause-effect pairs.

Besides, recent studies \cite{liu2020knowledge,cao2021knowledge} show that fusing two kinds of knowledge, namely the pretrained language models (PLMs) and the knowledge graphs (KGs), is crucial for ECE. 
However, they only use expensive manually annotated events to retrieve KGs, ignoring the useful knowledge of other elements in sentences. Furthermore, they simply encode the two types of knowledge separately, lacking effective knowledge fusion.

As shown in Table~\ref{tab:intro}, we summarize three key issues for ECE: 1)~\textbf{Complex Causality Extraction}, where multiple causal-effect pairs occur within a single sentence; 2)~\textbf{Subtask Interaction}, which involves modeling the mutual dependence between the two subtasks of ECE (i.e., EE and ECI); and 3)~\textbf{Knowledge Fusion}, which requires effectively fusing the knowledge from two kinds of modalities, i.e., PLMs and KGs.

\begin{table}[t]

\centering
\small
\caption{\label{tab:intro} Comparison of different methods on three key issues for ECE: Complex Causality Extraction (CCE), Subtask Interaction (SI), and Knowledge Fusion (KF).
}
\begin{tabular}{lccc}
\toprule
Method Type  &CCE& SI & KF\\
\midrule
Pipeline-Based \cite{kadowaki2019event}&\checkmark&$\times$&$\times$\\
Sequence Labeling-Based \cite{jinghang2020causal}&$\times$&\checkmark&$\times$\\
UniCE (ours) &\checkmark&\checkmark&\checkmark\\
\bottomrule
\end{tabular}

\end{table}

To address all three key issues simultaneously, we propose a unified ECE framework (\textbf{UniCE}), which consists of two multi-layer components: an event module for extracting events (EE) and a relation module for identifying causal relationships (ECI).
For an input sentence, the relation module first retrieves an initial background graph by taking the KG nodes mentioned in the input sentence and their few-hop neighbors in external KGs. The UniCE then performs the two subtasks of ECE in each layer, improving the prediction results layer by layer. The output of the last layer is used as the final prediction of UniCE.

For the \textbf{Complex Causality Extraction} issue, we simply employ separate decoders for EE and ECI, enabling the flexible extraction of multiple cause-effect pairs. We focus on how to address the other two issues without compromising the ability to extract complex causal relationships.
For the \textbf{Subtask Interaction} issue, we devise a subtask interaction mechanism: 1) we adopt stack-propagation to adapt the relation module to errors from the event module; 2) we employ a subtask information aggregator to transfer the prediction results of the relation module into the event~module.
For the \textbf{Knowledge Fusion} issue, we devise a knowledge fusion mechanism: 1) we retrieve knowledge related to each element in sentences, rather than manually annotated events; 2) we design an insertion induction module to dynamically connect the extracted events with retrieved knowledge, thus avoiding interference from irrelevant knowledge. 3) we employ a knowledge information aggregator to enable PLMs and KGs to fuse their information in the encoding process.
Both the subtask interaction and the knowledge fusion mechanism work across multiple layers, ensuring sufficient subtask interaction and knowledge fusion in a unified way.

Extensive experiments on three widely used data\-sets, EventStoryLine, SCIFI, and Causal-TimeBank, show our model achieves state-of-the-art (SOTA) performance and outperforms ChatGPT with a margin of at least 30\% F1 scores. Ablation studies demonstrate that our carefully devised subtask interaction and knowledge fusion mechanism can effectively improve the performance of UniCE. Besides, for the sentences with different numbers of causal pairs, our method achieves consistently better performance than baseline methods. Furthermore, external experiments show that our model can also effectively improve ChatGPT's ECE performance via in-context learning.

\section{Methodology}
\begin{figure*}[t]
  \centering
  \includegraphics[width=0.96\textwidth]{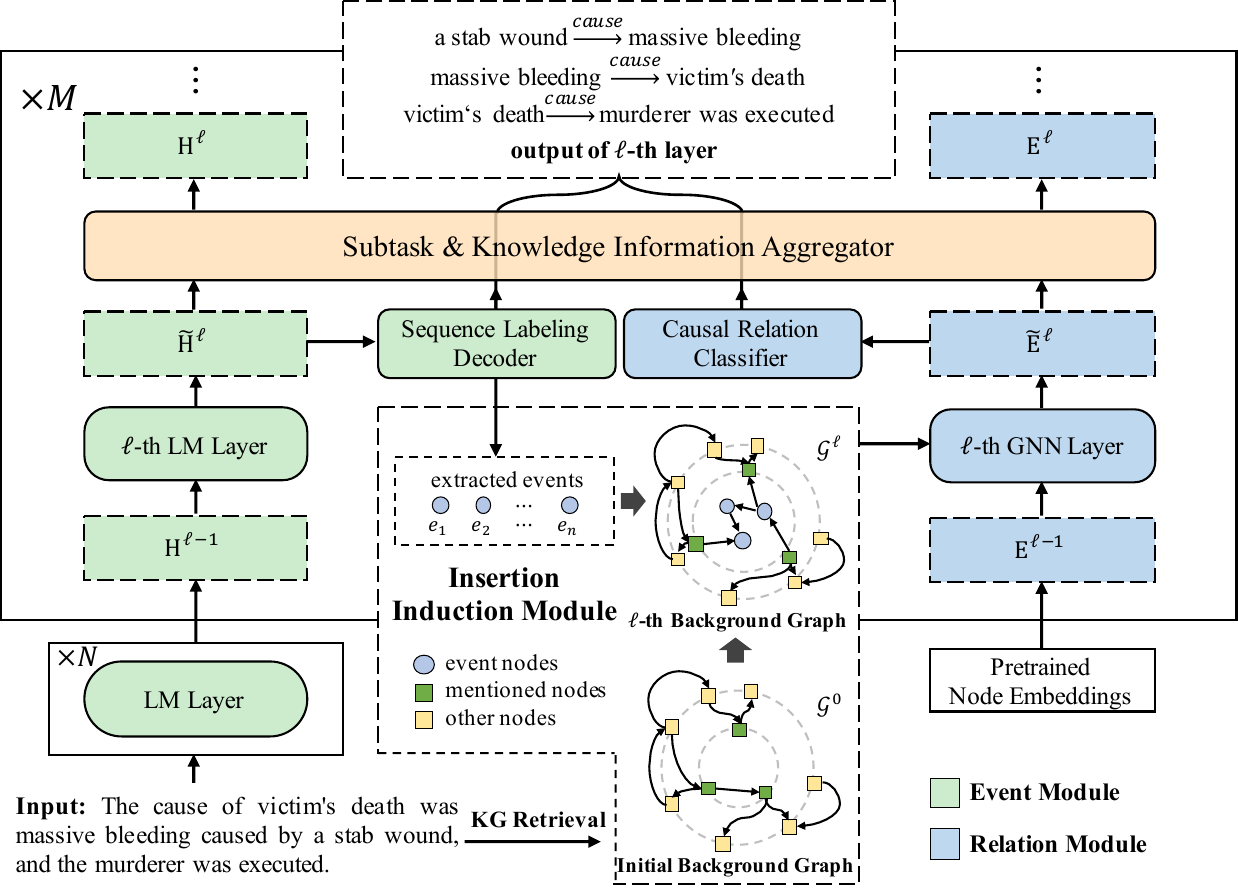}
  \caption{The illustration of the proposed unified ECE framework UniCE.}
  \label{fig:main}
\end{figure*}

\subsection{Overview of UniCE}
Figure~\ref{fig:main} shows the overview architecture of our proposed UniCE, which consists of two major modules: an event module with $N+M$ layers and a relation module with $M$ layers.
Given a sentence $S$, the first $N$ layers of the event module encode each token in $S$ using the first $N$ layers of PLMs. Besides, the relation module retrieves knowledge related to KG nodes mentioned in $S$ from external KGs to build an initial background graph $\mathcal{G}^{0}$.
In each of the following $M$ layers, the $\ell$-th layer of the event module first uses a PLM layer to update the representation of each token in $S$, and then adopts a sequence labeling decoder to extract events. These extracted events are then inserted into $\mathcal{G}^{0}$ by an insertion induction module to obtain the updated background graph $\mathcal{G}^{\ell}$.
After that, the relation module employs GNNs to encode and update the representations of nodes in $\mathcal{G}^{\ell}$ (including events extracted by the $\ell$-th event module layer), and then employs a classifier to judge the causal relationship between extracted events.
At the end of each subsequent $M$ layer, we feed the pre-fused representations of tokens in $S$ and nodes in $\mathcal{G}^{\ell}$ into two information aggregators to fuse information between the two subtasks and the two modalities of knowledge. The post-fused representations $\mathbf{H}^{\ell}$
and $\mathbf{E}^{\ell}$ are then used as the input of the next layer of our UniCE.
After the iterative fusion of $M$ layers, the output of the last layer is used as the final prediction of UniCE.

\subsection{Event Module} \label{event module}
The event module extracts events in $S$ that may be causally related to each other.

For the $\ell$-th layer of the event module, we first feed previous-layer-produced token representations $\mathbf{H}^{\ell-1}=\{\mathbf{h}_1^{\ell-1},\cdots,\mathbf{h}_n^{\ell-1}\}$ into a BERT layer to obtain the pre-fused token representations $\tilde{\mathbf{H}}^{\ell}=\{\tilde{\mathbf{h}}_1^{\ell},\cdots,\tilde{\mathbf{h}}_n^{\ell}\}$.
In the top $N$ layers, $\mathbf{H}^{\ell}$ is equal to $\tilde{\mathbf{H}}^{\ell}$. But in the next $M$ layers, $\mathbf{H}^{\ell}$ is computed by our two information aggregators.
Then, in each of the last $M$ layers, we extract events by feeding $\tilde{\mathbf{H}}^{\ell}$ into a Conditional Random Field (CRF) decoder:
\begin{equation}
    Y_e^{\ell}=\textrm{CRF}(\{\tilde{\mathbf{h}}_1^{\ell},\cdots,\tilde{\mathbf{h}}_n^{\ell}\}),
\end{equation}
where $Y_e^{\ell}=\{y_1^{\ell},\cdots,y_n^{\ell}\}$ is the predicted BIO tag sequence. Finally, we use the last token in each event as the pre-fused event context representation, denoted as $\tilde{\mathbf{H}}_{e}^{\ell}=\{\tilde{\mathbf{h}}_{e_1}^{\ell},\cdots,\tilde{\mathbf{h}}_{e_n}^{\ell}\}$.

\subsection{Relation Module}\label{relation module}
The relation module is designed to identify the causal relationship between the events extracted by the event module.
To adapt the relation module to errors from the event module, we utilizes $\tilde{\mathbf{H}}_{e}^{\ell}$ as inputs for the $\ell$-th relation module layer, rather than gold labeled events.
Besides, an insertion induction module is employed to dynamically connect the extracted events with retrieved knowledge, thus avoiding interference from irrelevant knowledge.

\subsubsection{Background Graph Construction:}
\textbf{\emph{1) Initialization.}}
Given an input sentence $S$, we first retrieve KGs to obtain an initial background graph $\mathcal{G}^{0}$.
Unlike previous ECI methods, we retrieve knowledge related to each element in $S$, rather than only gold-labeled events.
Specifically, we first retrieve external KGs to obtain KG nodes mentioned in $S$ as basic nodes $\mathcal{V}_{mention}$. Then, we add 2-hop neighbors of $\mathcal{V}_{mention}$ and any KG nodes that are in the shortest path (no more than 10 steps) between any pair of $\mathcal{V}_{mention}$ to get the set of related nodes $\mathcal{V}_{other}$.
The max number of nodes in $\mathcal{G}^{0}$ is set to 50.
Finally, we utilize all edges in KGs that connect any pairs of nodes in $\mathcal{V}_{other}$ and $\mathcal{V}_{mention}$ as edges in $\mathcal{G}^{0}$.
\textbf{\emph{2)~Dynamic Updating.}}\label{para:dynamic-update} In the $\ell$-th layer, the event module extracts events from $S$.
Then, we add the extracted events into the graph $\mathcal{G}^{0}$ as event nodes $\mathcal{V}_{event}^{\ell}$, and build weighted edges between nodes in $\mathcal{V}_{event}^{\ell} \cup \mathcal{V}_{mention}$ with our insertion induction module (detailed in \S\ref{insertion induction}.I). After that, we perform reasoning over the updated graph $\mathcal{G}^{\ell}$ (detailed in \S\ref{Reasoning over Graph}.R).

\subsubsection{Reasoning over Graph}\label{Reasoning over Graph}
In each of the $M$ layers, we employ GNNs on $\mathcal{G}^{\ell}$ to obtain the event node representations containing knowledge from both PLMs and external KGs.

For the $\ell$-th layer, we initialize the embeddings of nodes in $\mathcal{V}_{mention}$ and $\mathcal{V}_{other}$ with their post-fused node embeddings $\mathbf{E}^{\ell-1}$ produced by the previous layer.
And the embeddings of event nodes in $\mathcal{V}_{event}^{\ell}$ are initialized with $\tilde{\mathbf{H}}_{e}^{\ell}$. For the first layer, $\mathcal{V}_{mention}$ and $\mathcal{V}_{other}$ are initialized with pretrained embeddings provided by Feng et al. \cite{feng-etal-2020-scalable}.

In each of the $M$ layers, we feed $\mathbf{E}^{\ell-1}=\{\mathbf{e}_1^{\ell-1},\cdots,\mathbf{e}_{J}^{\ell-1}\}$ of nodes in $\mathcal{G}^{\ell}$ into the GNNs to obtain pre-fused node embeddings $\tilde{\mathbf{E}}^{\ell}=\{\tilde{\mathbf{e}}_1^{\ell},\cdots,\tilde{\mathbf{e}}_J^{\ell}\}$. And $J$ is the number of nodes in $\mathcal{G}^{\ell}$.
We follow previous work~\cite{yasunaga2021qagnn} to build the GNNs, though other GNN variants could also~be~used.

\subsubsection{Causal Relation Classifier}
In each of the $M$ layers, we employ a classifier to identify the causal relation between each extracted event pair $\left \langle i,j \right \rangle$ with their GNN-produced embeddings $\tilde{\mathbf{e}}_i^{\ell}$ and $\tilde{\mathbf{e}}_j^{\ell}$:
\begin{equation}
    y_{ij}^{\ell}=f_{R}([\tilde{\mathbf{e}}_i^{\ell};\tilde{\mathbf{e}}_j^{\ell}]),
\end{equation}
where $f_R$ is a 2-layer MLP with a softmax activation function, $y_{ij}^{\ell}$ indicates the causal relationship predicted by the $\ell$-th relation module layer.

\subsubsection{Insertion Induction} \label{insertion induction}
This module dynamically builds weighted edges between nodes in $\mathcal{V}_{event}^{\ell} \cup \mathcal{V}_{mention}$ to insert extracted events into $\mathcal{G}^{0}$.

We denote the edge weight matrix as $\mathbf{A}^{\ell}$. The value $\mathbf{A}_{ij}^{\ell}$ indicates the edge weight between nodes $\left \langle i,j \right \rangle$, where $i,j\notin \mathcal{V}_{other}$.
Because $\mathcal{V}_{event}^{\ell} \cup \mathcal{V}_{mention}$ are all in the same sentence, they are usually connected by some kind of syntactic dependency tree.
Thus, we use a variant of Kirchhoff's Matrix-Tree Theorem \cite{koo2007structured} to predict $\mathbf{A}^{\ell}$, which derives the link structure as a probabilistic expectation of possible dependency trees.

For the $\ell$-th layer, $\mathbf{e}_i^{\ell-1}$ indicates the input representation of the $i$-th node.
We first assign non-negative scores to the edges of $\mathbf{A}^{\ell}$:
\begin{equation}
    \mathbf{P}_{ij}=
\begin{cases}
\exp(f_a(\mathbf{e}_i^{\ell-1})^T\mathbf{W}_{c}f_b(\mathbf{e}_j^{\ell-1}))  & \text{ if } i\ne j   \\
0  & \text{ otherwise},
\end{cases}
\end{equation}
where $\mathbf{W}_c$ is a weight matrix, $f_a$ and $f_b$ are linear transformations with a tanh activation function, and $\mathbf{P}_{ij}$ is the score of the edge between the $i$-th and the $j$-th node. We then compute the root score $\mathbf{R}_i^r=\exp(\mathbf{W}_r\mathbf{e}_i^{\ell-1})$, which indicates the unnormalized probability of the $i$-th node to serve as the root of any dependency tree. After that, $\mathbf{A}^{\ell}$ is computed following the Matrix-Tree Theorem~\cite{koo2007structured} ($\delta$ is the Kronecker delta):
\begin{multicols}{2}
    \noindent
    \begin{equation}
        \mathbf{L}_{ij}=
\begin{cases}
\sum_{i'=1}^n\mathbf{P}_{i'j}  & \text{ if } i=j \\
-\mathbf{P}_{ij}  & \text{ otherwise},
\end{cases}
    \end{equation}
    \begin{equation}
        \hat{\mathbf{L}}_{ij}=
\begin{cases}
\mathbf{R}_j^r  & \text{ if } i=1 \\
\mathbf{L}_{ij}  & \text{ otherwise},
\end{cases}
    \end{equation}
\end{multicols}

\begin{equation}
        \begin{split}
        \mathbf{A}_{ij}^{\ell}=&(1-\delta _{1,j})\mathbf{P}_{ij}[\hat{\mathbf{L}}^{-1} ]_{ij}\\
        -&(1-\delta _{i,1})\mathbf{P}_{ij}[\hat{\mathbf{L}}^{-1} ]_{ji},
    \end{split}
    \end{equation}

\subsection{Subtask Information Aggregator}
\label{Event Representation Interaction}
To help EE benefit from the predictions of ECI, subtask information aggregator (T-aggregator) provides the ECI prediction information to EE in each of our last~M~layers.

In the $\ell$-th layer, our relation module feeds $\tilde{\mathbf{e}}_{e_i}^{\ell}$ into a simple classifier to identify causal relationships. Therefore, the ECI results are implicitly embedded into $\tilde{\mathbf{e}}_{e_i}^{\ell}$.
For the $i$-th extracted event, we first concatenate $\tilde{\mathbf{e}}_{e_i}^{\ell}$ output by GNNs and $\tilde{\mathbf{h}}_{e_i}^{\ell}$ output by PLMs, then feed them into T-aggregator to obtain post-fused representation $\mathbf{h}_{e_i}^{\ell}=\text{T-aggregator}([\tilde{\mathbf{h}}_{e_i}^{\ell};\tilde{\mathbf{e}}_{e_i}^{\ell}])$, where T-aggregator is a 2-layer MLP. Finally, we replace $\tilde{\mathbf{h}}_{e_i}^{\ell}$ in $\tilde{\mathbf{H}}^{\ell}$ with $\mathbf{h}_{e_i}^{\ell}$ to obtain $\mathbf{H}^{\ell}$, which is also the input of the next event module layer, thus providing ECI results for EE.

\subsection{Knowledge Information Aggregator}\label{Modality Fusion}

The knowledge information aggregator (K-aggregator) facilitates information fusion between the two kinds of knowledge, utilizing the mentioned KG nodes as the bridge.

In the $\ell$-th layer, the pre-fused context and knowledge embedding of the $i$-th KG node $a_i$ are $\tilde{\mathbf{h}}_{a_i}^{\ell}$ (output by PLMs) and $\tilde{\mathbf{e}}_{a_i}^{\ell}$ (output by GNNs), respectively.
$\tilde{\mathbf{h}}_{a_i}^{\ell}$ and $\tilde{\mathbf{e}}_{a_i}^{\ell}$ are concatenated and fed into K-aggregator: $[\mathbf{h}_{a_i}^{\ell};\mathbf{e}_{a_i}^{\ell}]=\text{K-aggregator}([\tilde{\mathbf{h}}_{a_i}^{\ell};\tilde{\mathbf{e}}_{a_i}^{\ell}])$, where K-aggregator is a 2-layer MLP. Then, we replace pre-fused node embeddings $\tilde{\mathbf{h}}_{a_i}^{\ell}$ and $\tilde{\mathbf{e}}_{a_i}^{\ell}$ in $\tilde{\mathbf{H}}^{\ell}$ and $\tilde{\mathbf{E}}^{\ell}$ with post-fused embeddings $\mathbf{h}_{a_i}^{\ell}$ and $\mathbf{e}_{a_i}^{\ell}$ to obtain $\mathbf{H}^{\ell}$
and $\mathbf{E}^{\ell}$, which are the input of the next layer of the event and the relation module, respectively.

\subsection{Multi-layer Learning \& Inference}

During training, the event module and the relation module perform sequence labeling and relation classification in each of their last $M$ layers.
We optimize the $\ell$-th layer of two modules by cross-entropy loss $\mathcal{L}_{e}^{\ell}$ and $\mathcal{L}_{r}^{\ell}$, respectively. The loss function of UniCE is defined as the average of the last $M$ layers:
\begin{equation}
    \mathcal{L}=\frac{1}{M} \sum_{\ell=1}^M \left (\mathcal{L}_{e}^{N+\ell} +\mathcal{L}_{r}^{\ell}\right).
\end{equation}
At inference time, we only use the last layer prediction to obtain extracted causal pairs.

\section{Experiments}

\subsection{Experimental Setup}

\subsubsection{Dataset and Evaluation Metrics}

Following previous ECE works \cite{liu2020knowledge,cao2021knowledge}, we employ three widely used datasets:
1) \textbf{EventStoryLine} v0.9 (ESC) \cite{caselli2017event}, which contains 258 documents, 5,334 events, and 1,770 causal event pairs;
2) \textbf{SCIFI} \cite{li2021causality}, which contains 5,236 sentences, and 1,866 causal event pairs. We remove duplicate negative examples;
3)~\textbf{Causal-TimeBank}, which contains 184 documents, 6,813 events, and 318 causal event pairs. Following previous works \cite{liu2020knowledge,cao2021knowledge}, we conduct 5-fold cross-validation on the ESC and SCIFI datasets, 10-fold cross-validation on the CTB dataset, respectively. We adopt the Precision (P), Recall (R), and F1-score (F1) as evaluation metrics.
All the results are the average of three independent experiments.

\subsubsection{Parameters Setting}\label{Parameters Settings}
We set $N$ and $M$ to 9 and 3, respectively.
We utilize a BERT-Base-Uncased \cite{kenton2019bert} architecture to implement our event module, which has 12-layers, 768-hiddens, and 12-heads.
The hidden size of other parameters is set to 200.
We choose ConceptNet as the external KG.
In each relation module layer, the number of GNN layers is set to 1.
The dropout of our model is set to 0.2.
We apply early stop and the Adam algorithm with a linear warmup schedule to optimize our model.
We set the batch size to 20 and use different learning rates for the LM encoder (lr=1e-5) and other parameters (lr=1e-4). Same as previous methods, we adopt a negative sampling strategy (rate=0.6) for the ESC and the CTB dataset.

\subsubsection{Baseline Methods}
\label{bm}

1) \textbf{Pipeline-based baseline methods:} these methods only reported their performances on the ECI task based on gold-labeled events. For a fair comparison, we train a BERT-CRF model to extract events as input for them.
- \emph{BERT-Pipeline}, first employs a BERT-CRF model to extract events, and then uses a BERT classifier to identify the causal relations.
- \emph{KMMG} \cite{liu2020knowledge}, a BERT-based model that utilizes external knowledge to enhance the representations of events.
- \emph{DPJL} \cite{shen-etal-2022-event}, a BERT-based model that incorporate information about causal cue words and the semantic relationship between events.
2)~\textbf{Sequence labeling-based baseline methods:}
- \emph{Nearest-BERT-CRF}, first uses a BERT-CRF model to extract causes and effects and then pairs each cause with the nearest effect in the sentence.
- \emph{SCITE} \cite{li2021causality}, first extracts causal events with the BERT-CRF model and then matches causes and effects into pairs with a set of handcraft rules.
3)~\textbf{Applying Joint Entity and Relation Extraction (JERE) methods to ECE:}
- \emph{CasRel} \cite{wei-etal-2020-novel}, a BERT-based model that utilizes the cascade framework for generic relation extraction.
- \emph{PRGC} \cite{zheng-etal-2021-prgc}, a BERT-based model that filters out low-confidence entity pairs to improve model performance.
- \emph{RFBFN} \cite{li-etal-2022-rfbfn}, a BERT-based model that incorporates semantic information of the target relationship.
4)~\textbf{ChatGPT based baseline methods:}
We conduct experiments with \emph{gpt-3.5-turbo} and employ a relaxed PRF calculation method. Specifically, a predicted causal-effect pair is considered correct if at least one token is shared between the predicted and the labeled cause, as well as between the predicted and the~labeled~effect.
- \emph{Zero-shot ChatGPT P1}, zero-shot predicting with the PROMPT 1 \footnote{Input: $\text{<input S>}$$\backslash$n Question: List the cause-effect pairs in the input sentence.$\backslash$n Answer:\_}.
- \emph{Zero-shot ChatGPT P2}, zero-shot predicting with the PROMPT 2 \footnote{Input: $\text{<input S>}$$\backslash$n 
Question: If there is a causal relationship between two events in the input sentence, extract the causal pair at the word level. If there are multiple causal pairs, add AND between them, otherwise answer None. For example: (accuse of) cause (death) AND (kill) cause (death)$\backslash$n 
Answer:\_}.
- \emph{Zero-shot ChatGPT P3}, zero-shot predicting with the PROMPT 3 \footnote{Input: $\text{<input S>}$$\backslash$n 
Question: Is there a token-level causal relationship in the sentence? If so, please extract it into this form: cause->effect. If there are multiple causal relationships, add AND between causal pairs, and display No if there is no causal relationship.$\backslash$n 
Answer:\_}.
- \emph{4-shot ChatGPT} or \emph{8-shot ChatGPT}, in-context learning with 4 or 8 demonstrations randomly selected from training sets, with PROMPT 3.
- \emph{UniCE-based ChatGPT}, with the same setup as in \emph{4-shot ChatGPT}, except that demonstrations are labeled with our UniCE and retrieval based on the semantic similarity of the questions.

\subsection{Experimental Results}

\begin{table*}[t]
\small
\centering
\caption{\label{tab:main}Experimental results of our model and the baselines.Bold denotes the best results. $\dagger$ denotes that BERT is used as the encoder. $\ddagger$ denotes the relaxed PRF describe in~\S\ref{bm}.
}
\begin{tabular}{lccccccccc}
\toprule
\multirow{2}{*}{\textbf{Methods}} & \multicolumn{3}{c}{\textbf{ESC}} & \multicolumn{3}{c}{\textbf{SCIFI}} & \multicolumn{3}{c}{\textbf{CTB}} \\
\cmidrule(lr){2-4}\cmidrule(lr){5-7} \cmidrule(lr){8-10}
% \cline{2-7}
 & \textbf{P}& \textbf{R}& \textbf{F1}& \textbf{P}& \textbf{R}& \textbf{F1} & \textbf{P}& \textbf{R}& \textbf{F1} \\
 \midrule
\textbf{Zero-shot ChatGPT P1}$^\ddagger$ & 0.0480 & 0.1625 & 0.0742 & 0.0915 & 0.2939 & 0.1395 & 0.0481 & 0.2919 & 0.0827\\
\textbf{Zero-shot ChatGPT P2}$^\ddagger$ & 0.0690 & 0.0795 & 0.0739 & 0.3237 & 0.2635 & 0.2905 & 0.0693 & 0.1577 & 0.0963\\
\textbf{Zero-shot ChatGPT P3}$^\ddagger$ & 0.1414 & 0.1043 & 0.1201 & 0.4522 & 0.3514 & 0.3954 & 0.0894 & 0.1678 & 0.1167\\
\textbf{4-shot ChatGPT}$^\ddagger$ & 0.1006 & 0.1810 & 0.1293 & 0.2668 & 0.4561 & 0.3367 & 0.0653 & 0.2181 & 0.1005\\
\textbf{8-shot ChatGPT}$^\ddagger$& 0.0981 & 0.1804 & 0.1271 & 0.2825 & 0.4696 & 0.3528 & 0.0631 & 0.2081 & 0.0968\\
\textbf{UniCE-based ChatGPT}$^\ddagger$ & 0.1937 & 0.2963 & 0.2342 & 0.3712 & 0.6182 & 0.4639 & 0.0598 & 0.1980 & 0.0918 \\
\midrule
\textbf{Nearest-BERT-CRF}$^\dagger$ & 0.4760 & 0.2756 & 0.3491 & \textbf{0.8729} & 0.5568 & 0.6799 & 0.3316 & 0.2493 & 0.2846\\
\textbf{SCITE}$^\dagger$ \cite{li2021causality} & 0.4547 & 0.3555 & 0.3990 & 0.8498 & 0.7259 & 0.7830 & 0.3033 & 0.3535 & 0.3265\\
\midrule
\textbf{CasRel}$^\dagger$ \cite{wei-etal-2020-novel} & 0.3929 & 0.3812 & 0.3870 & 0.7124 & 0.7766 & 0.7431 & 0.3649 & 0.3544 & 0.3596\\
\textbf{PRGC}$^\dagger$ \cite{zheng-etal-2021-prgc} & 0.4292 & 0.3839 & 0.4053& 0.7439 & 0.7703 & 0.7569& 0.3596 & 0.3267 & 0.3424\\
\textbf{RFBFN}$^\dagger$ \cite{li-etal-2022-rfbfn} & 0.4149 & 0.4091 & 0.4120& 0.7703 & 0.7630 & 0.7666& 0.3510 & 0.3844 & 0.3669\\
\midrule
\textbf{BERT-Pipeline}$^\dagger$ & 0.3990 & 0.3276 & 0.3598 & 0.7002 & 0.7742 & 0.7353& 0.2846 & 0.2530 & 0.2679\\
\textbf{KMMG}$^\dagger$ \cite{liu2020knowledge} & 0.3965 & 0.3899 & 0.3932 & 0.7750 & 0.7704 & 0.7727 & 0.3693 & 0.3972 & 0.3827\\
\textbf{DPJL}$^\dagger$ \cite{shen-etal-2022-event}& 0.4575 & 0.4071 & 0.4308 & 0.7477 & 0.7918 & 0.7691& 0.3839 & 0.4219 & 0.4020\\
\midrule
\textbf{UniCE}$^\dagger$ \textbf{(ours)}& \textbf{0.5419} & \textbf{0.4363} & \textbf{0.4834} & 0.8391 & \textbf{0.8236} & \textbf{0.8313} & \textbf{0.3923} & \textbf{0.4672} & \textbf{0.4265}\\
\bottomrule
\end{tabular}

\end{table*}

\subsubsection{Overall Performance}
Table~\ref{tab:main} shows the results on the ESC, SCIFI, and CTB datasets. We can find that:
Firstly, our model achieves SOTA performances.
These empirically shows that our proposed method can effectively capture cause-effect pairs in texts by facilitating the three major issues in ECE.
Secondly, although the JERE methods achieve acceptable performances, they cannot outperform the SOTA baseline methods. This is mainly because ECE is a knowledge-dependent reasoning task that also needs to model the mutual dependencies between the two ECE subtasks.
Besides, ChatGPT-based approaches perform poorly on ECE. This may be due to two reasons: 1) ChatGPT has limited training to structured output formats, which limits its performance on information extraction tasks \cite{wei2023zeroshot}. 2) ChatGPT may only partially understands the causal concept through causal trigger words (such as ``lead to'').
Moreover, utilizing the predictions generated by UniCE as demonstrations can effectively enhance the performance of ChatGPT. This is mainly because our UniCE, after fine-tuning, is more easily aligned with the ECE task objectives and adapted to structured output formats.

\subsubsection{Effect of Subtask Interaction and Knowledge Fusion}

\begin{table}[t]
  \centering
  \begin{minipage}{.45\textwidth}
    \small
\centering
\caption{\label{tab:fusion}Results with/without subtask interaction and kno\-wledge fusion mechanisms.
}
\begin{tabular}{lccc}
\toprule
\textbf{Model Setting} & \textbf{P} & \textbf{R} & \textbf{F1}\\
\midrule
\textbf{UniCE} & \textbf{0.5419} & \textbf{0.4363} & \textbf{0.4834} \\
\textbf{- w/o SI} & 0.4671 & 0.4107 & 0.4371 \\
\textbf{- w/o KF} & 0.4311 & 0.4191 & 0.4250 \\
\textbf{- w/o Both} & 0.3745 & 0.3706 & 0.3725 \\
\bottomrule
\end{tabular}
  \end{minipage}%
  \hfill
  \begin{minipage}{.45\textwidth}
    \small
\centering
\caption{\label{tab:taskfusion}Different directions of the interact\-ion between the two ECE subtasks.}
\begin{tabular}{lccc}
\toprule
\textbf{Model Setting} & \textbf{P} & \textbf{R} & \textbf{F1}\\
\midrule
\textbf{UniCE} &\textbf{0.5419} & \textbf{0.4363} & \textbf{0.4834} \\
\textbf{- w/o ECI to EE} & 0.5119 & 0.4253 & 0.4646 \\
\textbf{- w/o EE to ECI} & 0.4927 & 0.4281 & 0.4581 \\
\textbf{- w/o Both}  & 0.4671 & 0.4107 & 0.4371 \\
\bottomrule
\end{tabular}
  \end{minipage}
\end{table}

As shown in Table~\ref{tab:fusion}, we study the effectiveness of our devised subtask interaction and knowledge fusion mechanisms.
``w/o SI'' denotes that we feed gold-labeled events rather than extracted events into the relation module for training and remove the T-aggregator described in~\S\ref{Event Representation Interaction}. ``w/o KF'' denotes that there are no nodes in the initial background graph, and we remove the K-aggregator described in~\S\ref{Modality Fusion}. ``w/o Both'' denotes that we apply both of the above settings.
We can find that our model achieves lower F1 scores when our two mechanisms is removed. This indicates the effectiveness of our method.

\subsubsection{Effect of Subtask Interaction Directions}

As shown in Table~\ref{tab:taskfusion}, we analyze the different directions of the ECE subtask interaction.
``w/o ECI to EE'' denotes that we remove the T-aggregator.
``w/o EE to ECI'' denotes the relation module is trained with gold-labeled events.
``w/o Both'' denotes that we apply both of the above settings.
we find that the two interaction directions interactions can provide complementary benefits.

\subsubsection{Effect of Knowledge Fusion Components}

\begin{table}[t]
  \small
  \centering
  \begin{minipage}{.45\textwidth}
    \small
\centering
\caption{\label{tab:knwoledgefusion}Effect of different knowledge fusion components on the performance.}
\setlength{\tabcolsep}{3pt}
\begin{tabular}{lccc}
\toprule
\textbf{Model Setting} & \textbf{P} & \textbf{R} & \textbf{F1}\\
\midrule
\textbf{UniCE} & \textbf{0.541} & \textbf{0.436} & \textbf{0.483} \\
\textbf{- w/o PLM to KG} & 0.521 & 0.419 & 0.465\\
\textbf{- w/o KG to PLM} & 0.517 & 0.435 & 0.472\\
\textbf{\makecell[l]{- w/o Both}} & 0.485 & 0.414 & 0.447\\
\textbf{- w/o Insertion} & 0.478 & 0.424 & 0.449\\
\textbf{- w/o All} & 0.431 & 0.419 & 0.425 \\
\bottomrule
\end{tabular}
  \end{minipage}%
  \hfill
  \begin{minipage}{.45\textwidth}
    \small
\centering
\caption{\label{tab:gnn}Different methods for inserting extracted events into the background graph.}
\setlength{\tabcolsep}{5pt}
\begin{tabular}{lccc}
\toprule
\textbf{Method} & \textbf{P} & \textbf{R} & \textbf{F1}\\
\midrule
\textbf{No Link} & 0.478 & 0.424 & 0.449\\
\textbf{Span Match} & 0.484 & 0.429 & 0.455 \\
\textbf{Full Link} & 0.509 & 0.417 & 0.458 \\
\textbf{Dot-Product} & 0.519 & 0.414 & 0.461 \\
\midrule
\textbf{Our} &\textbf{0.541} & \textbf{0.436} & \textbf{0.483} \\
\bottomrule
\end{tabular}
  \end{minipage}
\end{table}

As shown in Table~\ref{tab:knwoledgefusion}, we analyze different knowledge fusion components for our method.
``w/o~PLM to KG'' denotes that in each layer, $\tilde{\mathbf{e}}_{a_i}^{\ell}$ are not used to update $\tilde{\mathbf{E}}^{\ell}$.
Similarly, ``w/o KG to PLM'' denotes $\tilde{\mathbf{h}}_{a_i}^{\ell}$ are not used to update $\tilde{\mathbf{H}}^{\ell}$. ``w/o~Both'' denotes that we remove the K-aggregator. ``w/o Insertion'' denotes that we insert extracted events into the graph without edges to other nodes. ``w/o All'' denotes that we apply all of them.
We find that both directions of knowledge fusion are important for the ECE task.

\subsubsection{Effect of Insertion Induction Module}

\begin{figure}[t]
  \centering
  \includegraphics[width=0.6\linewidth]{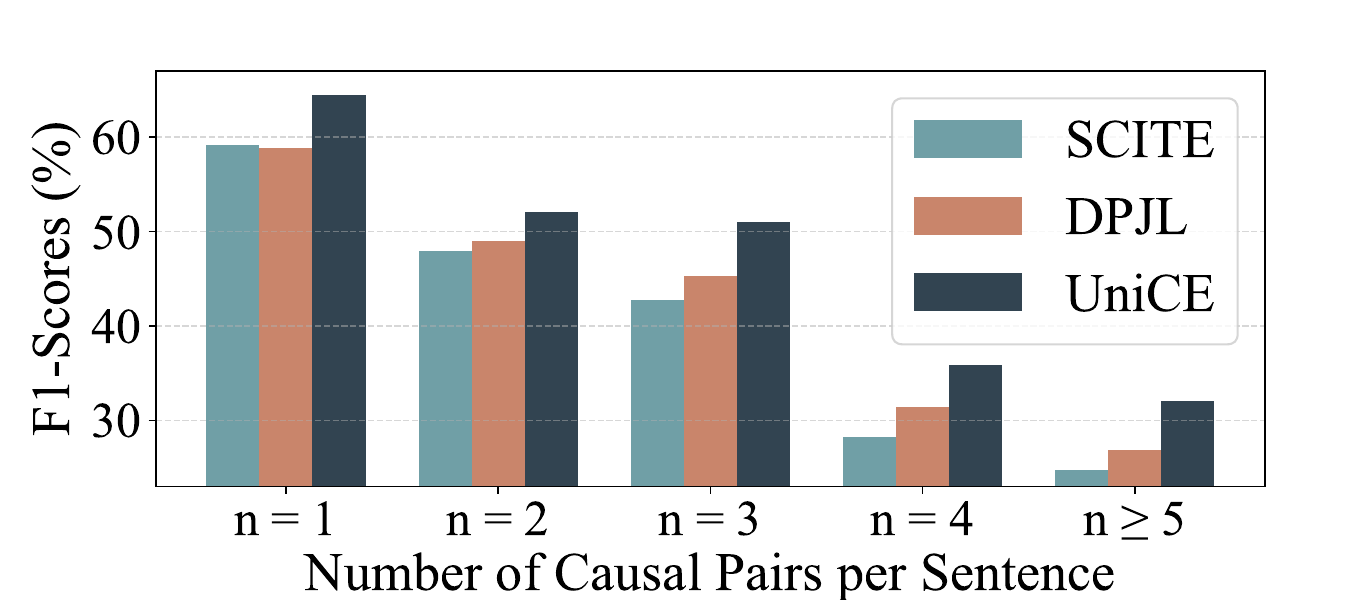}
  \caption{Experimental results on the ECE task with varying numbers of causal pairs per sentence.}
  \label{fig:pairnum}
\end{figure}

\begin{figure*}[t]
  \centering
  \includegraphics[width=0.9\linewidth]{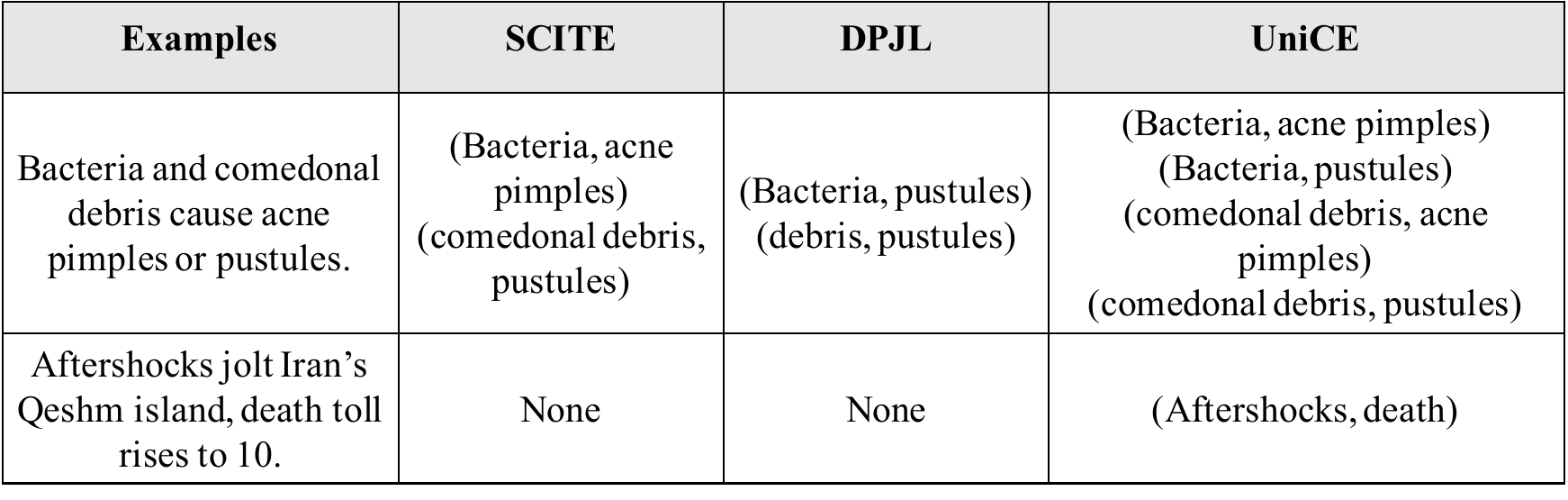}
  \caption{Case study of examples from SCIFI and EventStoryLine datasets.}
  
  \label{fig:case}
\end{figure*}

As shown in Table~\ref{tab:gnn}, we compare our insertion induction module with other four variants: 1) \textbf{No Link}, where we insert extracted events into the graph without edges to other nodes.
2) \textbf{Span Match}, if the token spans of an extracted event and a node are overlapped in the input sentence, we establish an edge.
3) \textbf{Full Link}, we establish edges between all nodes.
4) \textbf{Dot-Product}, we replace our insertion induction module with the Dot-Product Attention Mechanism.
We can observe that our model outperforms all four variants.

\subsubsection{Analysis of Complex Causal Extraction}

As shown in Figure~\ref{fig:pairnum}, we test models on sentences containing different numbers of cause-effect pairs. We can find that our method consistently outperforms best baseline DPJL and SCITE. This demonstrates that our framework could effectively deal with CCE issue.

\subsubsection{Case Study}

Figure~\ref{fig:case} shows two case study examples. In the first example, SCITE shows the weakness of dealing with the CCE issue. DPJL fails to extract the correct event and is not robust on wrong extracted event. In the second example, both SCITE and DPJL fail to identify the causal relation between ``Aftershocks'' and ``death''. While UniCE utilizes the knowledge \emph{(Aftershocks, CAUSE, collapse)} and \emph{(collapse, CAUSE, death)} to extract the causal pair correctly.

\section{Related Work}

\subsection{Event Causality Extraction}
Event Causality Extraction (ECE) aims to extract causal event pairs in texts.

Most recent methods address the ECE task with the pipeline framework.
Liu et al. \cite{liu2020knowledge} fed the knowledge related to candidate causal events from an external KG into a BERT encoder.
Zuo et al. \cite{zuo2020knowdis} proposed a data augmentation framework to the solve the data lacking problem of the ECE task.
Zuo et al. \cite{zuo-etal-2021-improving} leveraged external causal statements for event causality identification.
Liu et al. \cite{liu2023kept} incorporated background and relational information into the ECE model through prompt learning.
Shen et al. \cite{shen-etal-2022-event} proposed two prompt-based derivative tasks to utilize causal cue words and the relationship between events.
These methods only fuse the knowledge from PLMs and KGs in a separate and shallow manner.
To jointly learn the two subtasks, several studies \cite{jinghang2020causal,moghimifar2020domain} design sequence labeling-based methods for the ECE task. But they can only handle sentences with a single cause-effect pair.
Li et al. \cite{li2021causality} devised handcraft rules to pair causes and effects in the sentence, which cannot be generalized to other datasets.

Different from previous works, our framework can address all three key issues for ECE, i.e., complex causality extraction, subtask interaction, and~knowledge~fusion.

\subsection{Joint Entity and Relation Extraction}
The Joint Entity and Relation Extraction (JERE) task aims at extracting pairs of entities with semantic relations in texts.

Previous works utilize a cascade framework for joint extraction, which first extracted all possible subjects in texts, and then identified the corresponding objects for each subject \cite{wei-etal-2020-novel,zheng-etal-2021-prgc}.
In addition, some recent works first judge the semantic relationship between each token, and then transfer token-level relations into entity-level relations with handcraft rules \cite{wang-etal-2020-tplinker}.
Furthermore, several works \cite{zhao2021representation,li-etal-2022-rfbfn} introduced the semantics of relations as prior knowledge for the JERE task.

However, the dependence of EE on ECI is not trivial for the ECE task, which cannot be modeled by previous JERE approaches.
In addition, previous JERE works rarely study how to better introduce external knowledge into the extraction model.

\section{Speed Limitation}
Despite the effectiveness of our approach in causal extraction tasks, the incorporation of reasoning with knowledge graphs results in slower inference compared to baseline methods (for example, our inference speed is seven times slower than BERT). This indicates that our method might be challenging to apply directly in speed-sensitive applications. We believe this can be mitigated by introducing an intermediate scheduling module that can adaptively select the appropriate model based on the complexity of the input question. For instance, simple questions can be handled by the BERT baseline for extraction, while complex questions can utilize our proposed model. Moreover, there are still some design details in our approach that can be further optimized to enhance the runtime speed.

\section{Conclusion}

In this paper, we propose a multi-layer ECE method that is able to simultaneously address all three key issues for ECE, i.e., complex causality extraction, subtask interaction, and knowledge fusion.
Experimental results show that our model achieves consistently better performance than baseline methods on three widely used datasets. In particular, our model outperforms ChatGPT with a margin of at least 30\% F1-score. Moreover, experiments also show that our approach can effectively enhance the ECE performance of ChatGPT via in-context learning.

\section*{Acknowledgments}

We would like to thank the anonymous reviewers for their constructive comments, and gratefully acknowledge the support of National Key Laboratory of Information Systems Engineering (NO: 052022077), National Natural Science Foundation of China under Grants U22B2059 and 62176079, Natural Science Foundation of Heilongjiang Province under Grant YQ2022F005.

%
% ---- Bibliography ----
%
% BibTeX users should specify bibliography style 'splncs04'.
% References will then be sorted and formatted in the correct style.
%
\bibliographystyle{splncs04}
\bibliography{custom_short.bib}

\begin{thebibliography}{10}
\providecommand{\url}[1]{\texttt{#1}}
\providecommand{\urlprefix}{URL }
\providecommand{\doi}[1]{https://doi.org/#1}

\bibitem{cao2021knowledge}
Cao, P., Zuo, X., Chen, Y., Liu, K., Zhao, J., Chen, Y., Peng, W.: Knowledge-enriched event causality identification via latent structure induction networks. In: ACL (2021)

\bibitem{caselli2017event}
Caselli, T., Vossen, P.: The event {S}tory{L}ine corpus: A new benchmark for causal and temporal relation extraction. In: ACL Workshop. pp. 77--86 (2017)

\bibitem{kenton2019bert}
Devlin, J., Chang, M.W., Lee, K., Toutanova, K.: {BERT}: Pre-training of deep bidirectional transformers for language understanding. In: NAACL. pp. 4171--4186 (2019)

\bibitem{feng-etal-2020-scalable}
Feng, Y., Chen, X., Lin, B.Y., Wang, P., Yan, J., Ren, X.: Scalable multi-hop relational reasoning for knowledge-aware question answering. In: EMNLP. pp. 1295--1309 (2020)

\bibitem{fu2024jun}
Hao, F., Shanshan, L., Jiawei, L., Zhizheng, Z., Hui, Z.: Military event theme detection and extraction method. Command Informatipn System and Technology p.~015 (2024)

\bibitem{jinghang2020causal}
Jinghang, X., Wanli, Z., Shining, L., Ying, W.: Causal relation extraction based on graph attention networks. Journal of Computer Research and Development  \textbf{57}(1), ~159 (2020)

\bibitem{kadowaki2019event}
Kadowaki, K., Iida, R., Torisawa, K., Oh, J.H., Kloetzer, J.: Event causality recognition exploiting multiple annotators{'} judgments and background knowledge. In: EMNLP (2019)

\bibitem{koo2007structured}
Koo, T., Globerson, A., Carreras, X., Collins, M.: Structured prediction models via the matrix-tree theorem. In: EMNLP. pp. 141--150 (2007)

\bibitem{li2021causality}
Li, Z., Li, Q., Zou, X., Ren, J.: Causality extraction based on self-attentive bilstm-crf with transferred embeddings. Neurocomputing  \textbf{423},  207--219 (2021)

\bibitem{li-etal-2022-rfbfn}
Li, Z., Fu, L., Wang, X., Zhang, H., Zhou, C.: {RFBFN}: A relation-first blank filling network for joint relational triple extraction. In: ACL. pp. 10--20 (2022)

\bibitem{liu2020knowledge}
Liu, J., Chen, Y., Zhao, J.: Knowledge enhanced event causality identification with mention masking generalizations. In: Bessiere, C. (ed.) IJCAI. pp. 3608--3614 (2020)

\bibitem{liu2023kept}
Liu, J., Zhang, Z., Guo, Z., Jin, L., Li, X., Wei, K., Sun, X.: Kept: Knowledge enhanced prompt tuning for event causality identification. KBS  \textbf{259},  110064 (2023)

\bibitem{moghimifar2020domain}
Moghimifar, F., Haffari, G., Baktashmotlagh, M.: Domain adaptative causality encoder. In: Workshop of ALTA. pp. 1--10 (2020)

\bibitem{li2021jun}
Pengwei, L., Yazhao, L.: Event logic graph construction method for event profile. Command Informatipn System and Technology  \textbf{012},  54--60,69 (2021)

\bibitem{shen-etal-2022-event}
Shen, S., Zhou, H., Wu, T., Qi, G.: Event causality identification via derivative prompt joint learning. In: COLING. pp. 2288--2299 (2022)

\bibitem{wang-etal-2020-tplinker}
Wang, Y., Yu, B., Zhang, Y., Liu, T., Zhu, H., Sun, L.: {TPL}inker: Single-stage joint extraction of entities and relations through token pair linking. In: COLING. pp. 1572--1582 (2020)

\bibitem{wei2023zeroshot}
Wei, X., Cui, X., Cheng, N., Wang, X., Zhang, X., Huang, S., Xie, P., Xu, J., Chen, Y., Zhang, M., Jiang, Y., Han, W.: Zero-shot information extraction via chatting with chatgpt (2023)

\bibitem{wei-etal-2020-novel}
Wei, Z., Su, J., Wang, Y., Tian, Y., Chang, Y.: A novel cascade binary tagging framework for relational triple extraction. In: ACL. pp. 1476--1488 (Jul 2020)

\bibitem{yasunaga2021qagnn}
Yasunaga, M., Ren, H., Bosselut, A., Liang, P., Leskovec, J.: {QA}-{GNN}: Reasoning with language models and knowledge graphs for question answering. In: NAACL (2021)

\bibitem{zhao2021representation}
Zhao, K., Xu, H., Cheng, Y., Li, X., Gao, K.: Representation iterative fusion based on heterogeneous graph neural network for joint entity and relation extraction. KBS  \textbf{219} (2021)

\bibitem{zheng-etal-2021-prgc}
Zheng, H., Wen, R., Chen, X., Yang, Y., Zhang, Y., Zhang, Z., Zhang, N., Qin, B., Ming, X., Zheng, Y.: {PRGC}: Potential relation and global correspondence based joint relational triple extraction. In: ACL. pp. 6225--6235 (2021)

\bibitem{zuo-etal-2021-improving}
Zuo, X., Cao, P., Chen, Y., Liu, K., Zhao, J., Peng, W., Chen, Y.: Improving event causality identification via self-supervised representation learning on external causal statement. In: Findings of the ACL. pp. 2162--2172 (2021)

\bibitem{zuo2020knowdis}
Zuo, X., Chen, Y., Liu, K., Zhao, J.: Knowledge enhanced data augmentation for event causality detection via distant supervision. In: COLING. pp. 1544--1550 (2020)

\end{thebibliography}

\end{document}